\pdfoutput=1

\documentclass[11pt]{article}

\usepackage[final]{acl}

\usepackage{times}
\usepackage{latexsym}

\usepackage[T1]{fontenc}

\usepackage[utf8]{inputenc}

\usepackage{microtype}

\usepackage{inconsolata}

\usepackage{graphicx}

\usepackage{xspace}
\usepackage{longtable}
\usepackage{amssymb} 
\usepackage{pifont} 
\usepackage{float}
\usepackage{amsmath}
\usepackage{multirow}
\usepackage{caption}
\usepackage{booktabs}
\usepackage{graphicx}

\DeclareMathOperator*{\argmax}{arg\,max}

\usepackage{color}
\definecolor{blueVI}{RGB}{30, 136, 229}

\usepackage{graphicx}
\usepackage{tcolorbox}
\usepackage{listings}
\usepackage{xcolor}
\usepackage{hyperref}

\lstdefinestyle{mystyle}{
    backgroundcolor=\color{white},
    commentstyle=\color{teal!80!black},
    keywordstyle=\color{red},
    stringstyle=\color{blue},
    basicstyle=\ttfamily\footnotesize,
    breakatwhitespace=false,
    breaklines=true,         
    captionpos=b,                 
    keepspaces=true,                 
    showspaces=false,                 
    showstringspaces=false,           
    showtabs=false,                   
    tabsize=4
}

\newcommand{\ours}{\textsc{GreaTerPrompt}\xspace}

\title{\raisebox{-0.15\height}{\includegraphics[height=1em]{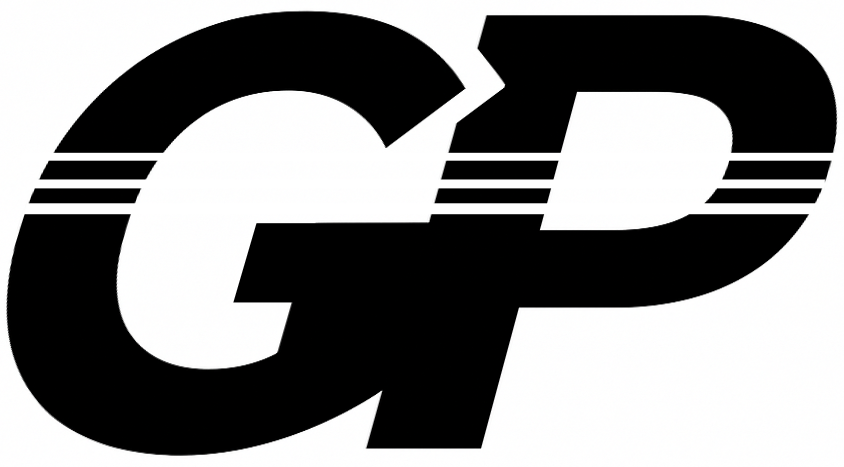}} \ours: A Unified, Customizable, and High-Performing Open-Source Toolkit for Prompt Optimization }

\author{Wenliang Zheng, Sarkar Snigdha Sarathi Das, Yusen Zhang, Rui Zhang \\
 Penn State University\\ 
 \texttt{\{wmz5132,sfd5525,yfz5488,rmz5227\}@psu.edu}}

\begin{document}
\maketitle
\begin{abstract}

LLMs have gained immense popularity among researchers and the general public for its impressive capabilities on a variety of tasks. Notably, the efficacy of LLMs remains significantly dependent on the quality and structure of the input prompts, making prompt design a critical factor for their performance. Recent advancements in automated prompt optimization have introduced diverse techniques that automatically enhance prompts to better align model outputs with user expectations. However, these methods often suffer from the lack of standardization and compatibility across different techniques, limited flexibility in customization, inconsistent performance across model scales, and they often exclusively rely on expensive proprietary LLM APIs. To fill in this gap, we introduce \ours, a novel framework that democratizes prompt optimization by unifying diverse methods under a unified, customizable API while delivering highly effective prompts for different tasks. Our framework flexibly accommodates various model scales by leveraging both text feedback-based optimization for larger LLMs and internal gradient-based optimization for smaller models to achieve powerful and precise prompt improvements. Moreover, we provide a user-friendly Web UI that ensures accessibility for non-expert users, enabling broader adoption and enhanced performance across various user groups and application scenarios. \ours is available at \url{https://github.com/psunlpgroup/GreaterPrompt} via GitHub, \href{https://pypi.org/project/greaterprompt/}{PyPI}, and web user interfaces. 

\begin{table*}[t!]
\centering
\renewcommand{\arraystretch}{1.3} 
\setlength{\tabcolsep}{4pt} 
\resizebox{\linewidth}{!}{%
    \begin{tabular}{cccccccccc}
    \toprule
    \textbf{Method} & \textbf{\begin{tabular}[c]{@{}c@{}}Text-based\\Optimization\end{tabular}} & \textbf{\begin{tabular}[c]{@{}c@{}}Gradient-based\\Optimization\end{tabular}} & \textbf{\begin{tabular}[c]{@{}c@{}}Zero-Shot\\Prompt\end{tabular}} & \textbf{\begin{tabular}[c]{@{}c@{}}Custom Metric\\Support\end{tabular}} & \textbf{Integration} & \textbf{\begin{tabular}[c]{@{}c@{}}Web UI\\Support\end{tabular}} & \textbf{\begin{tabular}[c]{@{}c@{}}Local Model\\Support\end{tabular}} & \textbf{\begin{tabular}[c]{@{}c@{}}Smaller Model\\Compatibility\end{tabular}} & \textbf{\begin{tabular}[c]{@{}c@{}}Larger Model\\Compatibility\end{tabular}} \\ 
    \midrule
    LangChain Promptim~\cite{promptim} & \checkmark & \texttimes & \checkmark & \checkmark & Library (Python API) & \texttimes & \checkmark & Low & High \\ 
    Stanford DsPy~\cite{khattab2024dspy} & \checkmark & \texttimes & Few-Shot & \checkmark & Library (Python) & \texttimes & \checkmark & Low & High \\
    AutoPrompt~\cite{2402.03099} & \checkmark & \texttimes & Few-Shot & \checkmark & Library (Python) & \texttimes & \texttimes & None & Limited \\
    Google Vertex Prompt Optimizer~\cite{gcd} & \checkmark & \texttimes & Few-Shot & \checkmark & Cloud & \texttimes & \texttimes & None & Proprietary Models Only \\
    AWS Bedrock Optimizer~\cite{bedrock} & Single Step rewrite & \texttimes & Few-Shot & \texttimes & Cloud & \texttimes & \texttimes & None & Proprietary Models Only \\
    Anthropic Claude Improver~\cite{anthropic} & LLM heuristic guided & \texttimes & \checkmark & \texttimes & Cloud & \texttimes & \texttimes & None & Proprietary Models Only \\
    Jina PromptPerfect~\cite{perfect} & LLM heuristic guided & \texttimes & \checkmark & \texttimes & Cloud & \checkmark & \texttimes & None & Limited \\
    \ours (Ours) & \checkmark & \checkmark & \checkmark & \checkmark & Library (Python) & \checkmark & \checkmark & High & High \\ \bottomrule
\end{tabular}
}
\caption{Comparison of different prompt optimization tools. While all existing methods rely on LLM feedback for \textbf{Text-Based Optimization}, \ours uniquely \textbf{also} supports \textbf{Gradient-based Optimization}, for more precise prompt tuning. Unlike methods requiring Few-Shot prompts, \ours deliver \textbf{Zero-Shot} optimized prompts. It also allows optimization for custom metrics—an option missing in many proprietary tools. Finally, \ours is the only method offering both an intuitive \textbf{Web-UI} and a \textbf{Python library}, with \textbf{high compatibility} across both small (locally deployed) and large (API-based) LLMs.}
\label{tab:comparison}
\end{table*}

\begin{figure*}[t!]
\centering
\includegraphics[width=\textwidth]{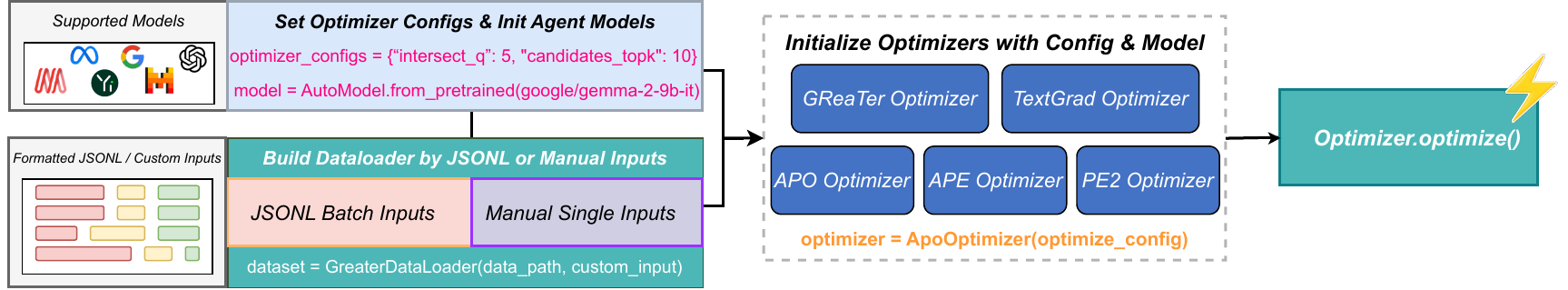}
\caption{Architecture Overview of \ours.}
\label{fig:architecture_overview}
\end{figure*}

\end{abstract}

\section{Introduction}

LLMs have demonstrated impressive capabilities across a wide variety of natural language tasks, establishing prompting as the primary means of communication between humans and machines~\cite{gu2023systematic}. However, despite the remarkable advances, LLMs remain highly sensitive to the prompt designs and formulations - that subtle variations in wordings of prompts can dramatically alter model outputs and affect performance~\cite{chatterjee-etal-2024-posix,zhuo-etal-2024-prosa}. This persistent prompt sensitivity implies that even the recent state-of-the-art LLMs do not entirely eliminate the need for careful prompt design, which traditionally relies on human expertise and iterative trial-and-error \cite{chen-etal-2024-prompt,wu2024ease}. In response, recent efforts have increasingly focused on developing automated prompt optimization methods~\cite{pryzant2023automatic,ye2023prompt,zhou2023largelanguagemodelshumanlevel,yuksekgonul2024textgrad,das2024greater}, systematically enhancing prompt quality and ensuring robust model performance without exhaustive manual tuning~\cite{wu2024ease,tang2025gpo}. 

However, as shown in Table~\ref{tab:comparison}, existing prompt optimization techniques and tools exhibit considerable variability in terms of usability, scope, and their performance often fluctuates inconsistently across different model scales. This variability and specialized nature often makes it challenging for non-expert users, who otherwise could derive substantial benefits from prompt optimization techniques while using LLMs. Moreover, existing prompt optimization methods rely on expensive proprietary LLM APIs, significantly undermining their affordability and privacy protection.

\begin{figure*}[t!]
    \centering
    \includegraphics[width=\textwidth]{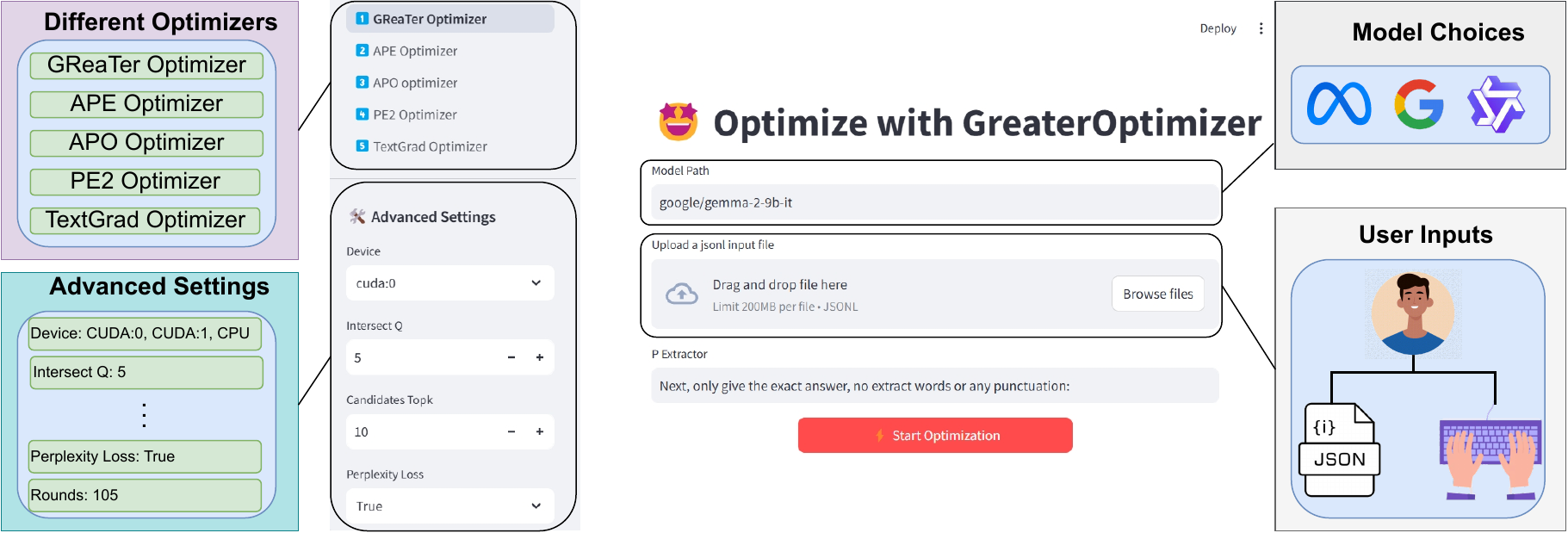}
    \caption{Screenshot of Web UI for \ours. Optimizer list is on the top left bar, bottom left bar is parameter settings for each optimizer. On the main area, there is a textbox for the model path input, and an area to upload user's prompt data. ``P Extractor'' is a system prompt for GReaTer optimizer to extract answer to calculate loss.}
    \label{fig:webui}
\end{figure*}

To bridge these gaps and encourage broad adoption of prompt optimization techniques, we introduce \ours, a novel framework designed to enhance accessibility, adaptability, and efficacy of prompt optimization. 
As shown in Figure~\ref{fig:architecture_overview}, \ours provides a streamlined workflow from inputs and model initialization to optimization execution, supporting flexible optimizer configurations that can be easily customized.
\ours is designed and implemented based on the following three principles:
\\\noindent \textbf{1) Methodological Perspective:} \ours unifies diverse prompt optimization methodologies under a cohesive implementation framework. Currently, \ours support five prominent prompt optimization techniques across two families based on model scales: i) Iterative Prompt Rewriting through LM feedback \cite{zhou2023largelanguagemodelshumanlevel,ye2023prompt,pryzant2023automatic,yuksekgonul2024textgrad}, and ii) Gradient based prompt optimization \cite{das2024greater}. This unification ensures users can leverage different types of LM feedback or gradient computations for optimizing LLM prompts.
\\\noindent\textbf{2) Model-Centric Perspective:} Larger, closed-source API-based LLMs like GPT~\cite{achiam2023gpt} and Gemini~\cite{team2024gemini} generally offer superior performance but are computationally expensive and require transmitting sensitive data externally; in contrast, smaller open-source LLMs like Llama 3~\cite{grattafiori2024llama} and Gemma 2~\cite{team2024gemma} 
provide cost-effective alternatives that better ensure data confidentiality. Recognizing the critical importance of model flexibility, \ours provides extensive support across both closed-source and open-source model families, ranging from compact, efficient models suitable for local deployment to large-scale models available via cloud APIs. By incorporating both gradient-based optimization techniques suitable for smaller models and feedback-driven optimization techniques for larger models, our framework ensures optimal performance irrespective of model choice and resource constraints.
\\\noindent\textbf{3) Integration Perspective:} \ours is designed with ease of use and integrability as key principles. To make it accessible for both expert and non-expert users, \ours offers both a Python package (a \href{https://github.com/psunlpgroup/GreaTer}{GitHub repository} and a \href{https://pypi.org/project/greaterprompt/}{pip} package) for simple incorporation into any existing pipeline and a user-friendly Web UI (Figure~\ref{fig:webui}) tailored for non-expert users. This dual interface design democratizes prompt optimization by enabling both expert and general users to benefit equally from state-of-the-art techniques. As shown in Table~\ref{tab:comparison}, \ours offers a comprehensive set of features compared to other libraries, supporting both text-based and gradient-based optimization while maintaining broad compatibility with smaller and larger models.

Overall, \ours combines flexible methodological support, extensive model compatibility, seamless integration, and comprehensive evaluation functionalities. 
\ours not only advances the state-of-the-art in prompt optimization but also makes these sophisticated techniques accessible to a broader audience, bridging the gap between research innovations and practical applications.
Our release include a GitHub repo \textbf{\url{https://github.com/psunlpgroup/GreaterPrompt}},
a PyPI installable package \textbf{\url{https://pypi.org/project/greaterprompt/}}, and a demo video \textbf{\url{https://youtu.be/aSLSnE17lBQ}}.

\section{Background} \label{sec:Background}
\subsection{Prompt Optimization Algorithms}
Given a task execution language model $f_{\text{LLM}}$, and a small representative task dataset, $\mathcal{D}_{task}=\{(x_1,y_1), \ldots (x_n,y_n)\}$, the goal of prompt optimization is to find a prompt $p^{*}$ such that:
\begin{equation}
    p^{*} = \argmax_{p}\sum_{(x,y) \in \mathcal{D}_{task}} m\left(f_{\text{LLM}}(x;p), y\right)
\end{equation}
where $f_{\text{LLM}}(x;p)$ is the output from $f_{\text{LLM}}$ upon channeling the input $x$ with the prompt $p$, and $m(\cdot)$ is the evaluation function for this task.

\noindent\textbf{Textual Feedback Based Prompt Optimization.} Prompt optimization methods based on textual feedback use an optimizer model $f_{\text{optimizer}}$ which is usually substantially larger and more expensive than $f_{\text{LLM}}$~\citep{zhou2023largelanguagemodelshumanlevel,ye2023prompt,pryzant2023automatic}. Conceptually, $f_{\text{optimizer}}\big(m(f_{\text{LLM}}(x;p),y)|(x,y)\in \mathcal{D}_{task}\big)$ drives the optimization process by assessing and providing feedback for refining the prompt. 

\noindent\textbf{Gradient Based Prompt Optimization.}
Traditional textual feedback-based prompt optimization relies heavily on the heuristic capabilities of large language models (LLMs) and often leads to poor performance when applied to smaller models. To overcome this, \ours introduces a stronger optimization signal in the form of loss gradients. The method begins by generating reasoning chains for task inputs using a small model. Then, it extracts final answer logits via a formatted prompt and computes loss. By backpropagating through this reasoning-informed output, optimizers will get a list of gradients with respect to candidate prompt tokens. These gradients are used to select optimal tokens, enabling efficient and effective prompt refinement—even for lightweight models.

 \subsection{Prompt Optimization Services/Libraries}
 Looking at Table 1, we can observe various prompt optimization methods currently available in the field. LangChain Promptim~\cite{promptim} offers text-based optimization with zero-shot capabilities and Python API integration. Stanford DsPy~\cite{khattab2022demonstrate}~\cite{khattab2024dspy} and AutoPrompt~\cite{2402.03099} provide similar text-based approaches with few-shot capabilities. Google Vertex Prompt Optimizer~\cite{gcd}, AWS Bedrock Optimizer~\cite{bedrock}, and Anthropic Claude Improver~\cite{anthropic} are cloud-based solutions with varying optimization techniques but limited model compatibility. Jina PromptPerfect~\cite{perfect} offers cloud integration with Web UI support but has limited model compatibility. \ours stands out by combining both text-based and gradient-based optimization approaches while supporting diverse integration options and high compatibility across model sizes. All of the existing services had some downsides and there was no unified way to use them until the introduction of \ours.

\subsection{Evaluation Metrics and Datasets for Prompt Optimization}

To evaluate the efficacy of the prompts produced by our library, in our experiments (Section~\ref{sec:high-performing}), we select two popular datasets for performance evaluation: BBH~\cite{suzgun2022challenging}, a diverse suite testing capabilities beyond current language models, and GSM8k~\cite{cobbe2021training} for mathematical reasoning assessment. All optimizers demonstrated performance improvements over the Zero-Shot Chain-of-Thought~\cite{kojima2022large}.

\section{\ours}
\ours is a unified (Section~\ref{sec:unified}), customizable (Section~\ref{sec:customizable}), and high-performing (Section~\ref{sec:high-performing}) library for prompt optimization.

\subsection{Unified Implementation}
\label{sec:unified}
\ours unifies the following five different prompt optimization methods under a single API. Even though existing methods already have released code, it is still challenging for beginner users for daily use. In our library, we build a unified data loading class. It supports both manual inputs by passing a list to the dataloader class and batch inputs by loading a jsonl file. With our data loader class, users could easily use all the supported optimizers with the same function calling, eliminating the need to initialize and optimize respectively by different methods.

\noindent \textbf{1) APE:} APE (Automatic Prompt Evolution)~\cite{zhou2023largelanguagemodelshumanlevel} is an optimization method that iteratively refines prompts for LLMs by automatically generating, evaluating, and evolving variations based on performance metrics. Inspired by evolutionary algorithms, it selects high-performing prompts, applies mutations, and repeats the process without requiring gradient-based tuning. This method reduces manual effort, enhances prompt effectiveness across tasks, and improves LLM performance in instruction following, reasoning, and factual accuracy.

\noindent \textbf{2) APO:} APO (Automated Prompt Optimization) \cite{pryzant2023automatic} is a technique that systematically refines prompts for large language models by leveraging iterative improvements. It evaluates multiple prompt variations, selects high-performing ones, and applies controlled modifications to enhance clarity, coherence, and effectiveness. Unlike manual tuning, APO automates the process using heuristic or model-driven feedback, ensuring better task-specific performance. This approach minimizes human effort, improves response reliability, and adapts to diverse use cases efficiently.

\noindent \textbf{3) GReaTer:} GReaTer \cite{das2024greater} is a novel prompt optimization method that enhances smaller language models by leveraging numerical gradients over task-specific reasoning instead of relying solely on textual feedback from large LLMs. Unlike existing techniques that depend on costly proprietary models like GPT-4, GReaTer enables self-optimization by computing loss gradients over generated reasoning steps. This approach improves prompt effectiveness and task performance in various reasoning benchmarks, achieving results comparable to or surpassing those of prompts optimized using massive LLMs.

\noindent \textbf{4) TextGrad:} TextGrad \cite{yuksekgonul2024textgrad} is a prompt optimization method that automates prompt refinement by leveraging textual feedback as a form of "textual gradient." Instead of using numerical loss gradients like GReaTer, TextGrad iteratively improves prompts based on feedback from a larger LLM, which critiques and suggests modifications to enhance task performance. This method relies on natural language evaluations of prompt effectiveness, guiding optimizations without requiring direct gradient computations. While effective in improving reasoning tasks, TextGrad can be computationally expensive and highly dependent on the quality of the feedback provided by the larger model.

\noindent \textbf{5) PE2:} PE2 (Prompt Engineering a Prompt Engineer) \cite{ye2023prompt} is a prompt optimization method that enhances prompts through meta-prompt engineering techniques. It iteratively refines prompts by analyzing model responses and leveraging structured feedback from large LLMs. PE2 systematically improves prompts by identifying patterns in successful completions and making targeted adjustments to optimize performance. While effective in improving reasoning and structured tasks, its reliance on external LLM-generated feedback can introduce variability, making optimization results dependent on the feedback model's quality.

\begin{table*}[t!]
\centering
\resizebox{\linewidth}{!}{%
\begin{tabular}{lccccc|c}
\toprule
\textbf{Optimizer} & \textbf{movie\_rec.} & \textbf{object\_count.} & \textbf{tracking\_five.} & \textbf{hyperbaton} & \textbf{causal} & \textbf{Average} \\
\midrule
ZS-CoT   & 47 & 67 & 49 & 68 & 51 & 56.4 \\
TextGrad \cite{yuksekgonul2024textgrad} & 48 & 80 & 55 & 66 & 42 & 58.2 \\
GReaTer \cite{das2024greater}  & \textbf{57} & \textbf{90} & \textbf{70} & \textbf{84} & \textbf{57} & \textbf{71.6} \\
\bottomrule
\end{tabular}

}
\caption{Performance in BBH tasks with GReaTer and TextGrad optimizers, with Llama3-8B-Instruction model. Here ZS-COT refers to: Zero-Shot Chain of Thought prompt i.e. "Let's think step by step".}
\label{tab:llama3 greater textgrad}
\vspace{0.3cm}  
\resizebox{\linewidth}{!}{%
\begin{tabular}{lccccc|c}
\toprule
\textbf{Optimizer} & \textbf{movie\_rec.} & \textbf{object\_count.} & \textbf{tracking\_five.} & \textbf{hyperbaton} & \textbf{causal} & \textbf{Average} \\
\midrule
ZS-CoT & 54 & 64 & 56 & 86 & 48 & 61.6 \\
APE \cite{zhou2023largelanguagemodelshumanlevel}   & 66 & 70 & 44 & 92 & 49 & 64.2 \\
APO \cite{pryzant2023automatic}   & 66 & 66 & 58 & \textbf{92} & \textbf{59} & 68.2 \\
PE2  \cite{ye2023prompt}  & \textbf{68} & \textbf{74} & \textbf{60} & 90 & 56 & \textbf{69.6} \\
\bottomrule
\end{tabular}

}
\caption{Performance in BBH tasks with APE, APO and PE2 optimized (with gpt-4-turbo) prompt used on gpt-3.5-turbo task model. Here ZS-COT refers to: Zero-Shot Chain of Thought prompt i.e. "Let's think step by step".}
\label{tab:gpt4 ape apo pe2}
\end{table*}

\subsection{User Customization}
\label{sec:customizable}
\ours allows users to choose task exemplar samples, evaluation functions, and model choices. 

\noindent\textbf{User-defined Task Examples.} User can upload their task examples consisting of input and output pairs in a JSON format, providing a demonstration of the target task which our library can use as oracle to produce the optimized prompts.

\noindent\textbf{Customized Task Evaluation Functions.} We found that cross entropy doesn't meet the needs of all tasks. To address this, we added support for custom loss functions in the GReaTer optimizer in our library. Users can define their own loss functions and pass them as a parameter to the model. The custom loss will then be used during backpropagation and gradient computation to help the optimizer choose better tokens.

\noindent\textbf{Flexible Model Choices.} Our library supports two types of model deployment: API-based and local. Both deployment modes are compatible with all model sizes. For the GReaTer method, users can choose smaller models like \texttt{meta-llama/Meta-Llama-3-8B-Instruct} for efficient optimization, or larger models like \texttt{meta-llama/Meta-Llama-3-70B-Instruct} to generate higher-quality token replacements. For the APO, APE, and PE2 methods, users can flexibly select GPT models ranging from the legacy \texttt{gpt-35-turbo} to the latest \texttt{gpt-4o} for evaluation and testing.

\subsection{High-Performing Prompts}
\label{sec:high-performing}
To demonstrate the performance of our five optimizers, we randomly sampled 5 subtasks from BBH for evaluation. For GReaTer and TextGrad optimizers, we choose \texttt{Llama3-8B-Instruction} as the optimization models, evaluation results can be found in Table~\ref{tab:llama3 greater textgrad} and Table~\ref{tab:gsm8ksmaller}. For APE, APO and PE2 optimizers, \texttt{gpt-4-turbo} as the optimization model and results can be found in Table~\ref{tab:gpt4 ape apo pe2} and Table~\ref{tab:gsm8klarger}.
The resulting prompts are in Table~\ref{tab:example}.

Based on the tables, the results demonstrate noteworthy performance differences between the various optimizers across the BBH subtasks. With the \texttt{Llama3-8B-Instruction} model, GReaTer achieves the highest average performance (71.6), outperforming both TextGrad (64.9) and the ZS-CoT baseline (56.4). For the \texttt{gpt-4-turbo} optimization model, PE2 shows the best overall performance (69.6), followed by APO (68.2), APE (64.2), and the ZS-CoT baseline (61.6). Notably, all optimizers demonstrate task-specific strengths, with hyperbaton being particularly receptive to optimization across both model types, while performance on causal reasoning remains more challenging. These results highlight the effectiveness of our optimizers across both large and small models on different tasks.



\section{Usage Examples}
\ours supports two ways of usage: Python package (Section~\ref{sec:python}) and web UI (Section~\ref{sec:web}). Our \href{https://youtu.be/aSLSnE17lBQ}{demo video} shows more details.

\subsection{Python Package}
\label{sec:python}
The following code snippets demonstrate a quick view of our library as a python package. 

\noindent\textbf{Data Loading.}
\ours supports two methods to build the dataloader. Users can either provide a jsonl file path to the predefined GreaterDataloader, which will automatically load batch inputs, or manually input samples. Each sample only needs to contain three mandatory keys: question, prompt, and answer.

\begin{tcolorbox}[colback=white, colframe=black, boxrule=1pt, arc=3pt]
\lstset{style=mystyle}
\begin{lstlisting}[language=Python]
# method 1: load jsonl file for batch inputs
dataset1 = GreaterDataloader(data_path=
"./data/boolean_expressions.jsonl")

# method 2: manually custom inputs
dataset2 = GreaterDataloader(custom_inputs=[
{"question": "((-1 + 2 + 9 * 5) - (-2 + -4 + -4 * -7)) =", 
"prompt": "Use logical reasoning and think step by step.", 
"answer": "24"},
{"question": "((-9 * -5 - 6 + -2) - (-8 - -6 * -3 * 1)) =",
"prompt": "Use logical reasoning and think step by step.",
"answer": "63"}])
\end{lstlisting}
\end{tcolorbox}

\noindent\textbf{Configs Initialization.}
\ours supports comprehensive and flexible configurations for each optimizer. Users can choose their desired model for optimization, either local or online. For the GReaTer optimizer, there are more advanced settings, and users can even customize their loss function to meet expectations for different tasks. For beginners, these fields can be left blank, as optimizers will initialize with default configurations.

\begin{tcolorbox}[colback=white, colframe=black, boxrule=1pt, arc=3pt]
\lstset{style=mystyle}
\begin{lstlisting}[language=Python]
optimize_config = {
"task_model": "openai_gpt35_turbo_instruct",
"optim_model": "openai_gpt4_turbo",
}
\end{lstlisting}
\end{tcolorbox}

\noindent\textbf{Optimizer Loading and Prompt Optimization.}
The initialization for optimizers is also very simple. If configurations have been defined, users can pass them to the optimizer as a parameter when initializing; otherwise, they can leave it blank. After that, users only need to call ~\texttt{.optimize()} for each optimizer and pass the predefined dataloader and initial prompt to the optimizer. After a brief waiting period, the optimizer will return either a single optimized prompt or a sequence of optimized prompts to the user. All processes are simple and highly integrated, requiring no specialized domain knowledge.

\begin{tcolorbox}[colback=white, colframe=black, boxrule=1pt, arc=3pt]
\lstset{style=mystyle}
\begin{lstlisting}[language=Python]
ape_optimizer = ApeOptimizer(optimize_config=optimize_config)
# config is optional
pe2_optimizer = Pe2Optimizer(optimize_config=optimize_config)
ape_result = ape_optimizer.optimize(dataloader1, p_init="think step by step")
pe2_result = pe2_optimizer.optimize(dataloader2, p_init="think step by step")
\end{lstlisting}
\end{tcolorbox}

\subsection{User Friendly Web Interface}
\label{sec:web}


A primary goal in building our library, \ours, is to democratize prompt optimization for both expert and non-expert users. Traditionally, as discussed in Section~\ref{sec:Background}, prompt optimization techniques have required a significant degree of technical expertise and coding proficiency, rendering them inaccessible to many end users. \ours addresses this barrier through a comprehensive and user-friendly web interface (see Figure~\ref{fig:webui}) that brings the power of automated prompt optimization to a broader audience. Through this interface, users only have to: \textbf{(i)} select from various prompt optimization methods; \textbf{(ii)} for API-based models, simply provide their model API key; \textbf{(iii)} for locally hosted models, specify the model path and select the target GPU. Finally, the interface exposes all core functionalities of the code-based library, including hyperparameter tuning, via intuitive controls such as steppers and dropdown menus—no coding required. We believe this UI-driven solution lowers the barrier to entry, making prompt optimization more accessible to users with varying levels of technical expertise.

\section{Conclusion}
\ours is a comprehensive open-source toolkit that supports features many other prompt optimization libraries lack. As shown in our comparison, it uniquely offers both iterative LLM-rewrite and gradient-guided optimization alongside zero-shot prompting and custom metrics. Its user-friendly web interface makes advanced prompt engineering accessible even to non-programmers, while supproting both smaller and larger models. We hope this tool will prove highly useful to a wide range of users, and that contributors will continue to enhance the platform by adding support for future prompt optimization techniques.

\newpage



\bibliography{custom}

\appendix

\section{GSM8K Results}

For mathematical reasoning, we compare the performance of different optimization algorithms with \ours on GSM8K \cite{cobbe2021training}. We evaluate the prompt performance on the dedicated test set of 1319 examples. Table \ref{tab:gsm8ksmaller} shows the performance of GreaTer \cite{das2024greater} and TextGrad \cite{yuksekgonul2024textgrad} with Llama-3-8B-Instruct optimized prompts.

\begin{table}[h]
\centering
\begin{tabular}{lc}
\toprule
\textbf{Optimizer} & \textbf{GSM8K} \\
\midrule
ZS-CoT   & 79.6 \\
TextGrad \cite{yuksekgonul2024textgrad} & 81.1 \\
GReaTer \cite{das2024greater}  & \textbf{82.6} \\
\bottomrule
\end{tabular}
\caption{GSM8K performance for ZS-CoT, TextGrad, and GReaTer with Llama-3-8B-Instruct}
\label{tab:gsm8ksmaller}
\end{table}
\noindent
\textbf{Larger Models} For prompt optimization performance comparison with larger model, we compare the performance in GSM8K with APE, APO, and PE2 as shown in Table \ref{tab:gsm8klarger}. Prompts are tested on Mistral-7B-Instruct-v0.2 as in \cite{ye2023prompt}.

\begin{table}[h]
\centering
\begin{tabular}{lc}
\toprule
\textbf{Optimizer} & \textbf{GSM8K} \\
\midrule
ZS-CoT & 48.1 \\
APE \cite{zhou2023largelanguagemodelshumanlevel}    & 49.7 \\
APO \cite{pryzant2023automatic}    & \textbf{51.0} \\
PE2 \cite{ye2023prompt}    & {50.5} \\
\bottomrule
\end{tabular}
\caption{GSM8K performance for ZS-CoT, APE, APO, and PE2, with gpt-4-turbo optimizer and Mistral-7B-Instruct-v0.2}
\label{tab:gsm8klarger}
\end{table}

\section{Optimized Prompts}

Table~\ref{tab:example} gives a list of optimized prompts for 5 randomly sampled BBH tasks by different prompt optimizers in \ours.

\clearpage
\onecolumn
\begin{longtable}{l|l|p{0.55\textwidth}}
\toprule
\textbf{Task} & \textbf{Method}  & \textbf{Prompt}   \\\midrule
\multirow{5}{*}{Movie Recommendation} & TextGrad & You will answer a reasoning question by explicitly connecting the events and outcomes, considering multiple perspectives and potential counterarguments.    \\\cmidrule{2-3}
& GREATER & Use causal diagram. The correct option asks whether the variable C has a causal relationship with D, based on changes in the probability P that C occurs given E.  \\\cmidrule{2-3}
& APE  & Approach each stage sequentially. \\\cmidrule{2-3}
& APO & Identify the direct cause of the outcome: was it the immediate action or condition without which the event wouldn't have occurred? \\\cmidrule{2-3}
& PE2  & Determine if the action was intentional and a contributing factor to the outcome. Answer 'Yes' if intentional and causative, 'No' otherwise.  \\\midrule
\multirow{5}{*}{Object Counting} & TextGrad & You will answer a reasoning question about counting objects. Think step by step, considering the context of the question and using it to inform your answer. Be explicit in your counting process, breaking it down.    \\\cmidrule{2-3}
& GREATER & Use only addition. Add step by step. Finally, give the correct answer.  \\\cmidrule{2-3}
& APE  & Let's continue by taking systematic, sequential steps.\\\cmidrule{2-3}
& APO & Let's think step by step.\\\cmidrule{2-3}
& PE2  & Let's identify and count the instances of the specified category of items mentioned, tallying multiples to determine their total quantity.  \\\midrule
\multirow{5}{*}{Tracking Shuffled Objects} & TextGrad & You will answer a reasoning question by providing a step-by-step breakdown of the process. Use vivid and descriptive language to describe the events, and make sure to highlight the key connections...    \\\cmidrule{2-3}
& GREATER & Use this process as an explanation stepwise for each step until you get to as given above Alice has got originaly the following as follows.  \\\cmidrule{2-3}
& APE  & We'll tackle this systematically, one stage at a time.\\\cmidrule{2-3}
& APO & Track ball swaps and position changes separately. List each swap, update positions and ball ownership after each, and determine final states for both.\\\cmidrule{2-3}
& PE2  & Let's carefully track each player's position swaps step by step to determine their final positions.  \\\midrule
\multirow{5}{*}{Hyperbaton} & TextGrad & You will answer a reasoning question. Think step by step. Provide explicit explanations for each step. Consider breaking down complex concepts into smaller, more manageable parts...    \\\cmidrule{2-3}
& GREATER & Use the reasoning and examples you would step. Finally give the actual correct answer.  \\\cmidrule{2-3}
& APE  & Approach this gradually, step by step\\\cmidrule{2-3}
& APO & Choose the sentence with adjectives in the correct order: opinion, size, age, shape, color, origin, material, purpose, noun."\\\cmidrule{2-3}
& PE2  & Let's think step by step, considering the standard order of adjectives in English: opinion, size, age, shape, color, origin, material, purpose.  \\\midrule
\multirow{5}{*}{Causal Judgment} & TextGrad & You will answer a reasoning question by explicitly connecting the events and outcomes, considering multiple perspectives and potential counterarguments...    \\\cmidrule{2-3}
& GREATER & Use causal diagram. The correct option ask about whether there the variable C of about whether a specific cause is sufficient. The answer a causal relationship between C to D if the probability P that C occurs given E changes.  \\\cmidrule{2-3}
& APE  & Approach each stage sequentially.\\\cmidrule{2-3}
& APO & Identify the direct cause of the outcome: was it the immediate action or condition without which the event wouldn't have occurred?\\\cmidrule{2-3}
& PE2  & Determine if the action was intentional and a contributing factor to the outcome. Answer 'Yes' if intentional and causative, 'No' otherwise.  \\
\bottomrule  
\caption{Results for 5 randomly sampled BBH tasks by 5 different optimizers}
\label{tab:example}
\end{longtable}

\end{document}